\documentclass[12pt,twoside]{article}
\usepackage{latexsym,amsmath,amssymb}
\def\eqvsp{}  \newdimen\paravsp  \paravsp=1.3ex
\def\,{\mskip 3mu} \def\>{\mskip 4mu plus 2mu minus 4mu} \def\;{\mskip 5mu plus 5mu} \def\!{\mskip-3mu}
\def\dispmuskip{\thinmuskip= 3mu plus 0mu minus 2mu \medmuskip=  4mu plus 2mu minus 2mu \thickmuskip=5mu plus 5mu minus 2mu}
\def\textmuskip{\thinmuskip= 0mu                    \medmuskip=  1mu plus 1mu minus 1mu \thickmuskip=2mu plus 3mu minus 1mu}
\textmuskip
\def\beq{\eqvsp\dispmuskip\begin{equation}}    \def\eeq{\eqvsp\end{equation}\textmuskip}
\def\beqn{\eqvsp\dispmuskip\begin{displaymath}}\def\eeqn{\eqvsp\end{displaymath}\textmuskip}
\def\bqa{\eqvsp\dispmuskip\begin{eqnarray}}    \def\eqa{\eqvsp\end{eqnarray}\textmuskip}
\def\bqan{\eqvsp\dispmuskip\begin{eqnarray*}}  \def\eqan{\eqvsp\end{eqnarray*}\textmuskip}

\newtheorem{theorem}{Theorem}
\newtheorem{corollary}[theorem]{Corollary}

\newtheorem{definition}[theorem]{Definition}

\newtheorem{proposition}[theorem]{Proposition}

\newtheorem{assumption}[theorem]{Assumption}
\newenvironment{keywords}{\centerline{\bf\small
Keywords}\begin{quote}\small}{\par\end{quote}\vskip 1ex}

\newtheorem{myexample}[theorem]{Example}

\def\paradot#1{\vspace{\paravsp plus 0.5\paravsp minus 0.5\paravsp}\noindent{\bf\boldmath{#1.}}}
\def\paranodot#1{\vspace{\paravsp plus 0.5\paravsp minus 0.5\paravsp}\noindent{\bf\boldmath{#1}}}
\def\req#1{(\ref{#1})}

\def\fr#1#2{{\textstyle{#1\over#2}}}
\def\frs#1#2{{^{#1}\!/\!_{#2}}}
\def\SetR{I\!\!R}
\def\SetN{I\!\!N}

\def\qmbox#1{{\quad\mbox{#1}\quad}}
\def\e{{\rm e}}                        

\def\E{{\bf E}}                         
\def\P{{\rm P}}                         

\def\v{\boldsymbol}
\def\trp{{\!\top\!}}

\def\a{\alpha}

\def\g{\gamma}

\def\t{\theta}

\def\T{\Theta}
\def\H{{\cal H}}
\def\vx{\v x}
\def\X{{\cal X}}
\def\LossP{{\text{Loss}}}
\def\LossT{{\text{L}\widetilde{\text{oss}}}}
\def\PHIP{\text{PHI}}
\def\PHIT{\smash{\widetilde{\text{PHI}}}}
\def\MAP{{\text{MAP}}}
\def\IMAP{{\text{IMAP}}}
\def\ML{{\text{ML}}}
\def\eoe{\hspace*{\fill} $\diamondsuit\quad$\\} 

\begin{document}

\title{
\vskip 2mm\bf\Large\hrule height5pt \vskip 4mm
Predictive Hypothesis Identification
\vskip 4mm \hrule height2pt}
\author{{\bf Marcus Hutter}\\[3mm]
\normalsize RSISE$\,$@$\,$ANU and SML$\,$@$\,$NICTA \\
\normalsize Canberra, ACT, 0200, Australia \\
\normalsize \texttt{marcus@hutter1.net \ \  www.hutter1.net}
}
\date{8 September 2008}
\maketitle
\vspace*{-3ex}

\begin{abstract}
While statistics focusses on hypothesis testing and on estimating
(properties of) the true sampling distribution, in machine
learning the performance of learning algorithms on future data is
the primary issue.
In this paper we bridge the gap with a general principle (PHI) that
identifies hypotheses with best predictive performance. This includes
predictive point and interval estimation, simple and composite
hypothesis testing, (mixture) model selection, and others as special
cases.
For concrete instantiations we will recover well-known methods,
variations thereof, and new ones.
PHI nicely justifies, reconciles, and blends (a reparametrization
invariant variation of) MAP, ML, MDL, and moment estimation.
One particular feature of PHI is that it can genuinely
deal with nested hypotheses.
\def\contentsname{\centering\normalsize Contents}\\
{\parskip=-2.7ex\tableofcontents}
\end{abstract}

\vspace*{-2ex}

\newpage
\section{Introduction}\label{secIntro}

Consider data $D$ sampled from some distribution $p(D|\t)$
with unknown $\t\in\Omega$.
The likelihood function or the posterior contain the complete
statistical information of the sample. Often this information needs
to be summarized or simplified for various reasons
(comprehensibility, communication, storage, computational
efficiency, mathematical tractability, etc.). Parameter estimation,
hypothesis testing, and model (complexity) selection can all be
regarded as ways of summarizing this information, albeit in
different ways or context.
The posterior might either be summarized by a single point
$\T=\{\t\}$ (e.g.\ ML or MAP or mean or stochastic model selection),
or by a convex set $\T\subseteq\Omega$ (e.g.\ confidence or credible
interval), or by a finite set of points $\T=\{\t_1,...,\t_l\}$
(mixture models) or a sample of points (particle filtering), or by
the mean and covariance matrix (Gaussian approximation), or by more
general density estimation, or in a few other ways
\cite{Borovkov:98,Bishop:06}. I have roughly sorted the methods in
increasing order of complexity. This paper concentrates on set
estimation, which includes (multiple) point estimation and
hypothesis testing as special cases, henceforth jointly referred to
as ``{\it hypothesis identification}'' (this nomenclature seems
uncharged and naturally includes what we will do: estimation and
testing of simple and complex hypotheses but not density
estimation).
We will briefly comment on generalizations beyond set estimation at
the end.

\paradot{Desirable properties}
There are many desirable properties any hypothesis identification
principle ideally should satisfy. It should
\begin{itemize}\parskip=0ex\parsep=0ex\itemsep=0ex
\item lead to good predictions (that's what models are ultimately for),
\item be broadly applicable,
\item be analytically and computationally tractable,
\item be defined and make sense also for non-i.i.d.\ and non-stationary data,
\item be reparametrization and representation invariant,
\item work for simple and composite hypotheses,
\item work for classes containing nested and overlapping hypotheses,
\item work in the estimation, testing, and model selection regime,
\item reduce in special cases (approximately) to existing other methods.
\end{itemize}
Here we concentrate on the first item, and will show that the resulting
principle nicely satisfies many of the other items.

\vspace{1ex}\paradot{The main idea}
We address the problem of identifying hypotheses (parameters/models)
with good {\em predictive performance} head on. If $\t_0$ is the
true parameter, then $p(\vx|\t_0)$ is obviously the best prediction
of the $m$ future observations $\vx$. If we don't know $\t_0$ but
have prior belief $p(\t)$ about its distribution, the predictive
distribution $p(\vx|D)$ based on the past $n$ observations $D$
(which averages the likelihood $p(\vx|\t)$ over $\t$ with posterior
weight $p(\t|D)$) is by definition the best Bayesian predictor Often
we cannot use full Bayes (for reasons discussed above) but predict
with hypothesis $H=\{\t\in\T\}$, i.e.\ use $p(\vx|\T)$ as
prediction. The closer $p(\vx|\T)$ is to $p(\vx|D)$ or
$p(\vx|\t_0,D)$\footnote{So far we tacitly assumed that given
$\t_0$, $\vx$ is independent $D$. For non-i.i.d.\ data this is
generally not the case, hence the appearance of $D$.} the better is
$H$'s prediction (by definition), where we can measure closeness
with some distance function $d$. Since $\vx$ and $\t_0$ are (assumed
to be) unknown, we have to sum or average over them.

\begin{definition}[Predictive Loss]\label{defLossP}
The predictive \LossP / $\LossT$
of $\T$ given $D$ based on distance $d$ for $m$ future
observations is
\bqa\label{LPT}
  \LossP_d^m(\T,D) &:=& \int d(p(\vx|\T),p(\vx|D))d\vx \\ \nonumber
  \LossT_d^m(\T,D) &:=& \iint d(p(\vx|\T),p(\vx|\t,D))\,p(\t|D)d\vx\,d\t
\eqa
\end{definition}
{\em Predictive hypothesis identification} (PHI) minimizes the
losses w.r.t.\ some {\em hypothesis class} $\H$. Our formulation is
general enough to cover point and interval estimation, simple and
composite hypothesis testing, (mixture) model (complexity)
selection, and others.

\paradot{(Un)related work}
The general idea of inference by maximizing predictive performance
is not new \cite{Geisser:93}.
Indeed, in the context of model (complexity) selection it is
prevalent in machine learning and implemented primarily by empirical
cross validation procedures and variations thereof
\cite{Zucchini:00} or by minimizing test and/or train set
(generalization) bounds; see \cite{Langford:02} and references
therein.
There are also a number of statistics papers on predictive
inference; see \cite{Geisser:93} for an overview and older
references, and \cite{Barbieri:04,Mukho:05} for newer references.
Most of them deal with distribution free methods based on some form
of cross-validation discrepancy measure, and often focus on model
selection. A notable exception is MLPD \cite{Lejeune:82}, which
maximizes the predictive likelihood including future observations.
The full decision-theoretic setup in which a decision based on $D$
leads to a loss depending on $x$, and minimizing the expected loss,
has been studied extensively \cite{Borovkov:98,Hutter:04uaibook},
but scarcely in the context of hypothesis identification.
On the natural progression of
estimation$\rightarrow$prediction$\rightarrow$action, approximating
the predictive distribution by minimizing \req{LPT} lies between
traditional parameter estimation and optimal decision making.
Formulation \req{LPT} is quite natural but I haven't seen it
elsewhere. Indeed, besides ideological similarities the papers above
bear no resemblance to this work.

\paradot{Contents}
The main purpose of this paper is to investigate the predictive
losses above and in particular their minima, i.e.\ the best
predictor in $\H$.
Section \ref{secPrelim} introduces notation, global assumptions, and
illustrates PHI on a simple example. This also shows a shortcoming
of MAP and ML esimtation.
Section \ref{secPHIP} formally states PHI, possible distance
and loss functions, their minima,
In Section \ref{secExact}, I study exact properties of PHI:
invariances, sufficient statistics, and equivalences.
Sections \ref{secmggn} investigates the limit $m\to\infty$
in which PHI can be related to MAP and ML.
Section \ref{secmlln} derives large sample approximations
$n\to\infty$ for which PHI reduces to sequential moment fitting (SMF).
The results are subsequently used for Offline PHI.
Section \ref{secDisc} contains summary, outlook and conclusions.
Throughout the paper, the Bernoulli example will illustrate the
general results.

\paranodot{The main aim}
of this paper is to introduce and motivate PHI,
demonstrate how it can deal with the difficult problem of selecting
composite and nested hypotheses, and show how PHI reduces to known
principles in certain regimes.
The latter provides additional justification and support of previous
principles, and clarifies their range of applicability. In general,
the treatment is exemplary, not exhaustive.

\section{Preliminaries}\label{secPrelim}

\paradot{Setup}
Let $D\equiv D_n\equiv (x_1,...,x_n)\equiv x_{1:n}\in\X^n$ be the
observed {\em sample} with {\em observations} $x_i\in\X$ from some
measurable space $\X$, e.g.\ $\SetR^{d'}$ or $\SetN$ or a subset
thereof. Similarly let $\vx\equiv(x_{n+1},...,x_{n+m})\equiv
x_{n+1:n+m}\in\X^m$ be potential {\em future observations}. We
assume that $D$ and $\vx$ are sampled from some {\em probability
distribution} $\P[\cdot|\t]$, where $\t\in\Omega$ is some unknown
parameter. We do {\em not} assume independence of the
$x_{i\in\SetN}$ unless otherwise stated. For simplicity of
exposition we assume that the {\em densities} $p(D|\t)$ w.r.t.\ the
default (Lebesgue or counting) measure ($\int d\lambda$, $\sum_x$,
written both henceforth as $\int dx$) exist.

\paradot{Bayes}
Similarly, we assume a prior distribution $\P[\T]$ with density
$p(\t)$ over parameters. From {\em prior} $p(\t)$ and {\em
likelihood} $p(D|\t)$ we can compute the {\em posterior}
$p(\t|D)=p(D|\t)p(\t)/p(D)$, where normalizer $p(D)=\int
p(D|\t)p(\t)d\t$. The full Bayesian approach uses parameter
averaging for prediction
\beqn
  p(\vx|D) \;=\; \int p(\vx|\t,D)p(\t|D)d\t \;=\; {p(D,\vx)\over p(D)}
\eeqn
the so-called {\em predictive distribution} (or more precisely
predictive density), which can be regarded as the gold standard for
prediction (and there are plenty of results justifying this
\cite{Barron:93,Hutter:04uaibook}).

\paradot{Composite likelihood}
Let $H_\t$ be the {\em simple hypothesis} that $\vx$ is sampled
from $p(\vx|\t)$ and $H_\T$ the {\em composite hypothesis} that
$\vx$ is ``sampled'' from $p(\vx|\T)$, where $\T\subseteq\Omega$.
In the Bayesian framework, the ``{\em composite likelihood}''
$p(\vx|\T)$ is actually well defined (for measurable $\T$ with
$\P[\T]>0$) as an averaged likelihood
\beqn
  p(\vx|\T) \;= \int p(\vx|\t)p(\t|\T)d\t,\quad
  \qmbox{where} p(\t|\T)={p(\t)\over\P[\T]} \mbox{ for $\t\in\T$ and $0$ else}.
\eeqn

\paradot{MAP and ML}
Let $\H$ be the (finite, countable, continuous, complete, or else)
class of hypotheses $H_\T$ (or $\T$ for short) from which the
``best'' one shall be selected. Each $\T\in\H$ is assumed to be a
measurable subset of $\Omega$.
The {\em maximum a posteriori} (MAP) estimator is defined as $\t^\MAP =
\arg\max_{\t\in\H} p(\t|D)$ if $\H$ contains only simple hypotheses
and $\T^\MAP = \arg\max_{\T\in\H} \P[\T|D]$ in the general case.
The composite {\em maximum likelihood} estimator is defined as $\T^\ML =
\arg\max_{\T\in\H} p(D|\T)$, which reduces to ordinary ML for simple
hypotheses.

In order not to further clutter up the text with too much
mathematical gibberish, we make the following global assumptions
during informal discussions:

\newtheorem{globalassumptions}[theorem]{Global Assumption}
\begin{globalassumptions}
Wherever necessary, we assume that sets, spaces, and functions are
measurable, densities exist w.r.t.\ some (Lebesgue or counting) base
measure, observed events have non-zero probability, or densities
conditioned on probability zero events are appropriately defined, in
which case statements might hold with probability 1 only. Functions
and densities are sufficiently often (continuously) differentiable,
and integrals exist and exchange.
\end{globalassumptions}

\paradot{Bernoulli Example}
Consider a binary $\X=\{0,1\}$ i.i.d.\ process 
$p(D|\t)=\t^{n_1}(1-\t)^{n_0}$ with bias $\t\in[0,1]=\Omega$, and
$n_1=x_1+...+x_n=n-n_0$ the number of observed 1s. Let us assume a
uniform uniform prior $p(\t)=1$. Here but not generally in later
continuations of the example we also assume $n_0=n_1$.
Consider hypothesis class $\H=\{H_f,H_v\}$ containing simple
hypothesis $H_f=\{\t=\fr12\}$ meaning ``fair'' and composite vacuous
alternative $H_v=\Omega$ meaning ``don't know''. It is easy to see
that
%
\beqn
  p(D|H_v) = p(D) = {n_1!n_0!\over(n\!+\!1)!} \;<\; 2^{-n} = p(D|H_f)
  \qmbox{for $n>1\quad$ (and = else)}
\eeqn
hence $\T^\ML=H_f$, i.e.\ {\bf ML} always suggests a fair coin however
weak the evidence is. On the other hand, $\P[H_f|D]=0<1=\P[H_v|D]$,
i.e.\ {\bf MAP} never suggests a fair coin however strong the evidence is.

Now consider {\bf PHI}. Let $m_1=x_{n+1}+...x_{n+m}=m-m_0$ be the number
of future 1s. The probabilities of $\vx$ given $H_f$, $H_v$, and $D$
are, respectively
\beq\label{pxEx}
  p(\vx|H_f) = 2^{-m},\quad
  p(\vx|H_v) = {m_1!m_0!\over(m+1)!},\quad
  p(\vx|D) = {(m_1\!+\!n_1)!(n_0\!+\!m_0)!\over(n\!+\!m\!+\!1)!}{(n\!+\!1)!\over n_1!n_0!}
\eeq

For {\boldmath$m=1$} we get $p(1|H_f)=\fr12=p(1|H_v)$, so when concerned
with predicting only one bit, both hypotheses are equally good.
More generally, for an interval $\T=[a,b]$, compare
$p(1|\T)=\bar\t:=\fr12(a+b)$ to the full Bayesian prediction
$p(1|D)={n_1+1\over n+2}$ (Laplace's rule). Hence if $\H$ is a class
of interval hypotheses,
then PHI chooses the $\T\in\H$ whose midpoint $\bar\t$ is closest to
Laplace's rule, which is reasonable. %
The size of the interval doesn't matter, since $p(x_{n+1}|\T)$
is independent of it.

Things start to change for {\boldmath$m=2$}. The following table lists
$p(\vx|D)$ for some $D$, together with $p(\vx|H_f)$ and
$p(\vx|H_v)$, and their prediction error Err$(H):=\LossP_1^2(H,D)$
for $d(p,q)=|p-q|$ in \req{LPT}
\beqn\small
\def\frs#1#2{#1/#2}
\begin{array}{c|ccc||c|c}
  p(\vx|D) & \vx=00 & \vx=01|10 & \vx=11 & \mbox{Err}(H_f)\gtrless\mbox{Err}(H_v) & \mbox{Conclusion} \\\hline
  D=\{\}   & \frs13 & \frs13    & \frs13 & \frs13\quad>\quad\; 0\;    & \mbox{don't know} \\
  D=01     &\frs3{10}&\frs4{10} &\frs3{10}& \frs15\quad>\quad \frs2{15}& \mbox{don't know} \\
  D=0101   & \frs27 & \frs37    & \frs27 & \frs17\quad<\quad \frs4{21}& \mbox{fair}       \\
  D=(01)^\infty & \frs14 & \frs12 & \frs14 & \;0\;\quad<\quad \frs13  & \mbox{fair}       \\\hline
  p(\vx|H_f)& \frs14 & \frs12 & \frs14 \\
  p(\vx|H_v)& \frs13 & \frs13 & \frs13 \\
\end{array}
\eeqn
The last column contains the identified best predictive hypothesis.
For four or more observations, PHI says ``fair'', otherwise ``don't
know''.

Using \req{pxEx} or our later results, one can show
more generally that PHI chooses ``fair'' for {\boldmath$n\gg m$} and
``don't know'' for {\boldmath$m\gg n$}.
\eoe

\paradot{MAP versus ML versus PHI}
The conclusions of the example generalize: For $\T_1\subseteq\T_2$,
we have $\P[\T_1|D]\leq\P[\T_2|D]$, i.e.\ MAP always chooses the
less specific hypothesis $H_{\T_2}$. On the other hand, we have
$p(D|\t^\ML)\geq p(D|\T)$, since the maximum can never be smaller
than an average, i.e.\ composite ML prefers the maximally specific
hypothesis.
So interestingly, although MAP and ML give identical answers for
uniform prior on simple hypotheses, their naive extension to
composite hypotheses is diametral. While MAP is risk averse finding
a likely true model of low predictive power, composite ML risks an
(over)precise prediction.
Sure, there are ways to make MAP and ML work for nested hypotheses.
The Bernoulli example has also shown that PHI's answer depends not
only on the past data size $n$ but also on the future data size $m$.
Indeed, if we make only few predictions based on a lot of data
($m\ll n$), a point estimation ($H_f$) is typically sufficient,
since there will not be enough future observations to detect any
discrepancy.
On the other hand, if $m\gg n$, selecting a vacuous model ($H_v$)
that ignores past data is better than selecting a potentially wrong
parameter, since there is plenty of future data to learn from.
This is exactly the behavior PHI exhibited in the example.

\section{Predictive Hypothesis Identification Principle}\label{secPHIP}

We already have defined the predictive loss functions in
\req{LPT}. We now formally state our predictive hypothesis
identification (PHI) principle, discuss possible distances $d$, and
major prediction scenarios related to the choice of $m$.

\paradot{Distance functions}
Throughout this work we assume that $d$ is continuous and
zero if and only if both arguments coincide. Some popular
distances are:
the (f) $f$-divergence $d(p,q)=f(p/q)q$ for convex $f$ with $f(1)=0$, %
the ($\alpha$) $\alpha$-distance $f(t)=|t^\alpha-1|^{1/\alpha}$, %
the (1) absolute deviation $d(p,q)=|p-q|$ ($\alpha=1$), %
the (h) Hellinger distance $d(p,q)=(\sqrt p-\sqrt q)^2$ ($\alpha=\fr12$), %
the (c) chi-square distance $f(t)=(t-1)^2$, %
the (k) KL-divergence $f(t)=t\ln t$, and
the (r) reverse KL-divergence $f(t)=-\ln t$. %
The only distance considered here that is not an $f$ divergence %
is the (2) squared distance $d(p,q)=(p-q)^2$.
The $f$-divergence is particularly interesting, since it contains
most of the standard distances and makes \LossP\ representation
invariant (RI).

\begin{definition}[Predictive hypothesis identification (PHI)]
The best ($\widetilde{\smash{\text{best}}}$) predictive hypothesis in
$\H$ given $D$ is defined as
\beqn
  \hat\T_d^m \;:= \arg\min_{\T\in\H}\LossP_d^m(\T,D) \qquad
  (\,\tilde\T_d^m := \arg\min_{\T\in\H}\LossT_d^m(\T,D)\,)
\eeqn
The \PHIP\ ($\PHIT$) principle states to predict $\vx$ with
probability $p(\vx|\hat\T_d^m)$ ($p(\vx|\tilde\T_d^m)$), which we
call $\PHIP_d^m$ ($\PHIT_d^m$) prediction.
\end{definition}

\paradot{Prediction modes}
There exist a few distinct prediction scenarios and modes. Here are
prototypes of the presumably most important ones:
{\em\bfseries Infinite batch:} Assume we summarize our data $D$ by a
model/hypothesis $\T\in\H$. The model is henceforth used as
background knowledge for predicting and learning from further
observations essentially indefinitely. This corresponds to
$m\to\infty$.
{\em\bfseries Finite batch:} Assume the scenario above, but
terminate after $m$ predictions for whatever reason. This
corresponds to a finite $m$ (often large).
{\em\bfseries Offline:} The selected model $\T$ is used for
predicting $x_{k+1}$ for $k=n,...,n+m-1$ separately with
$p(x_{k+1}|\T)$ without further learning from $x_{n+1}...x_k$ taking
place. This corresponds to repeated $m=1$ with common $\T$:
$\LossP_d^{1m}(\T,D) := \E[\sum_{k=n}^{n+m-1}\LossP_d^1(\T,D_k)|D]$.
{\em\bfseries Online:} At every step $k=n,...,n+m-1$ we determine a (good)
hypothesis $\T_k$ from $\H$ based on past data $D_k$, and use it
only once for predicting $x_{k+1}$. Then for $k+1$ we select a new
hypothesis etc. This corresponds to repeated $m=1$ with different
$\T$: $\LossP=\sum_{k=n}^{n+m-1}\LossP_d^1(\T_k,D_k)$.

The above list is not exhaustive. Other prediction scenarios are
definitely possible. In all prediction scenarios above we can use
$\LossT$ instead of $\LossP$ equally well. Since all time steps $k$
in Online PHI are completely independent, online PHI reduces to
1-Batch PHI, hence will not be discussed any further.

\section{Exact Properties of PHI}\label{secExact}

\paradot{Reparametrization and representation invariance (RI)}
An important sanity check of any statistical procedure is its
behavior under reparametrization $\t\leadsto\vartheta=g(\t)$
\cite{Kass:96} and/or when changing the representation of
observations $x_i\leadsto y_i=h(x_i)$ \cite{Walley:96}, where $g$
and $h$ are bijections. If the parametrization/representation is
judged irrelevant to the problem, any inference should also be
independent of it. MAP and ML are both representation invariant, but
(for point estimation) only ML is reparametrization invariant.

\begin{proposition}[Invariance of \LossP]\label{propI}
$\LossP_d^m(\T,D)$ and $\LossT_d^m(\T,D)$ are invariant under
reparametrization of $\Omega$. If distance $d$ is an $f$-divergence,
then they are also independent of the representation of the
observation space $\X$. For continuous $\X$, the transformations are
assumed to be continuously differentiable.
\end{proposition}

RI for $\LossP_f^m$ is obvious, but will see later some
interesting consequences. Any exact inference or any specialized
form of PHI$_f$ will inherit RI. Similarly for approximations, as long as
they do not break RI. For instance, PHI$_h$ will lead to an interesting
RI variation of MAP.

\paradot{Sufficient statistic}
For large $m$, the integral in Definition \ref{defLossP} is
prohibitive. Many models (the whole exponential family) possess  a
sufficient statistic which allows us to reduce the integral over
$\X^m$ to an integral over the sufficient statistic.
Let
\beq\label{defSS}
  \mbox{$T:\X^m\to\SetR^{d'}$ be a sufficient statistic, i.e. $p(\vx|T(\vx),\t)=p(\vx|T(\vx))\;\forall \vx,\t$}
\eeq
which implies that there exist functions $g$ and $h$ such that
the likelihood factorizes into
\beq\label{phg}
  p(\vx|\t) \;=\; h(\vx)g(T(\vx)|\t)
\eeq
The proof is trivial for discrete $\X$ (choose
$h(\vx)=p(\vx|T(\vx))$ and $g(t|\t)=p(t|\t):=\P[T(\vx)=t|\t]$) and
follows from Fisher's factorization theorem for continuous $\X$.
Let $A$ be an event that is independent $\vx$ given $\t$.
Then multiplying \req{phg} by $p(\t|A)$ and integrating over $\t$
yields
\bqa\label{phgA}
  p(\vx|A) &=& \int p(\vx|\t)p(\t|A)d\t \;=\; h(\vx)g(T(\vx)|A),\qmbox{where}
\\ \label{defgtA}
  g(t|A) &:=& \int g(t|\t)p(\t|A)d\t
\eqa
For some $\beta\in\SetR$ let (non-probability) measure
$\mu_\beta[B]:=\int_{\{\vx:T(\vx)\in B\}} h(\vx)^\beta d\vx$
($B\subseteq\SetR^{d'}$) have density $h_\beta(t)$ ($t\in\SetR^{d'}$)
w.r.t.\ to (Lebesgue or counting) base measure $dt$
($\int dt=\sum_t$ in the discrete case).
Informally,
\beq\label{defhb}
  h_\beta(t) \;:=\; \int h(\vx)^\beta\delta(T(\vx)-t)d\vx
\eeq
where $\delta$ is the Dirac delta for continuous $\X$ (or the
Kronecker delta for countable $\X$, i.e.\
$\int d\vx\,\delta(T(\vx)-t)=\sum_{\vx:T(\vx)=t}$).

\begin{theorem}[PHI for sufficient statistic]\label{thmSS}
Let $T(\vx)$ be a sufficient statistic \req{defSS} for $\t$ and
assume $\vx$ is independent $D$ given $\t$, i.e.\
$p(\vx|\t,D)=p(\vx|\t)$. Then
\bqan
  \LossP_d^m(\T,D) &=& \int d(g(t|\T),g(t|D))h_\beta(t)dt \\
  \LossT_d^m(\T,D) &=& \int d(g(t|\T),g(t|\t))h_\beta(t)p(\t|D)dtd\t
\eqan
holds (where $g$ and $h_\beta$ have been defined in \req{phg},
\req{defgtA}, and \req{defhb}), provided one (or both) of the
following conditions hold:
(i) distance $d$ scales with a power $\beta\in\SetR$, i.e.\
$d(\sigma p,\sigma q)=\sigma^\beta d(p,q)$ for $\sigma>0$, or
(ii) any distance $d$, but $h(\vx)\equiv 1$ in \req{phg}.
One can choose $g(t|\cdot)=p(t|\cdot)$, the probability density of
$t$, in which case $h_1(t)\equiv 1$.
\end{theorem}

All distances defined in Section \ref{secPHIP} satisfy
$(i)$, the $f$-divergences all with $\beta=1$ and the square loss
with $\beta=2$.
The independence assumption is rather strong. In practice, usually
it only holds for some $n$ if it holds for all $n$. Independence of
$x_{n+1:n+m}$ from $D_n$ given $\t$ for all $n$ can only be
satisfied for independent (not necessarily identically distributed)
$x_{i\in\SetN}$.

\begin{theorem}[Equivalence of $\PHIP_{2|r}^m$ and $\PHIT_{2|r}^m$]\label{thmP2eqT2}\label{thmPreqTr}
For square distance ($d\widehat=2$) and RKL distance
($d\widehat=r$), $\LossP_d^m(\T,D)$ differs from $\LossT_d^m(\T,D)$
only by an additive constant c$_d^m(D)$ independent of $\T$, hence \PHIP\ and
$\PHIT$ select the same hypotheses $\hat\T_2^m=\tilde\T_2^m$ and
$\hat\T_r^m=\tilde\T_r^m$.
\end{theorem}

\paradot{Bernoulli Example}
Let us continue with our Bernoulli example with uniform prior.
$T(\vx)=x_1+...+x_m=m_1=t\in\{0,...,m\}$ is a sufficient statistic.
Since $\X=\{0,1\}$ is discrete, $\int dt=\sum_{t=0}^m$ and $\int
dx=\sum_{\vx\in\X^m}$. In \req{phg} we can choose $g(t|\t) =
p(\vx|\t) = \t^t(1-\t)^{m-t}$ which implies $h(\vx)\equiv 1$ and
$h_\beta(t)=\sum_{\vx:T(\vx)=t} 1={\textstyle\binom{m}{t}}$. From
definition \req{phgA} we see that $g(t|D)=p(\vx|D)$ whose expression
can be found in \req{pxEx}. For RKL-distance, Theorem \ref{thmSS}
now yields $\LossP_r^m(\T|D) = \sum_{t=1}^m h_\beta(t)
g(t|D)\ln{g(t|D)\over g(t|\T)}$. For a point hypothesis $\T=\{\t\}$
this evaluates to a constant minus $m[{n_1+1\over n+2}\ln\t +
{n_1+1\over n+2}\ln(1\!-\!\t)]$, which is minimized for
$\t={n_1+1\over n+2}$. Therefore the best predictive point
$\hat\t_r={n_1+1\over n+2} = \tilde\t_r$ = Laplace rule, where we
have used Theorem \ref{thmPreqTr} in the third equality. \eoe

\section{PHI for $\infty$-Batch}\label{secmggn}

In this section we will study PHI for large $m$, or more precisely,
the $m\gg n$ regime. No assumption is made on the data size $n$,
i.e.\ the results are exact for any $n$ (small or large) in the
limit $m\to\infty$. For simplicity and partly by necessity we assume
that the $x_{i\in\SetN}$ are i.i.d. (lifting the ``identical'' is
possible). Throughout this section we make the following
assumptions.

\begin{assumption}\label{assmggn}
Let $x_{i\in\SetN}$ be independent and identically distributed,
$\Omega\subseteq\SetR^d$, the likelihood density $p(x_i|\t)$ twice
continuously differentiable w.r.t.\ $\t$,
and the boundary of $\T$ has zero prior probability.
\end{assumption}

We further define $x:=x_i$ (any $i$) and the partial derivative
$\partial:=\partial/\partial\t=
(\partial/\partial\t_1,...,\partial/\partial\t_d)^\trp=(\partial_1,...,\partial_d)^\trp$.
The (two representations of the) {\em Fisher information matrix} of
$p(x|\t)$
\beq\label{FI}
  I_1(\t) \;=\; \E[(\partial\ln p(x|\t))(\partial\ln p(x|\t))^\trp|\t]
  \;=\; - \int (\partial\partial^\trp\ln p(x|\t)) p(x|\t)dx
\eeq
will play a crucial role in this Section. It also occurs in {\em
Jeffrey's prior},
\beq\label{JP}
  p_J(\t) \;:=\; \sqrt{\det I_1(\t)}/J,\qquad
  J \;:=\; \int \sqrt{\det I_1(\t)}d\t
\eeq
a popular reparametrization invariant (objective) reference prior
(when it exists) \cite{Kass:96}. We call the determinant (det) of
$I_1(\t)$, {\em Fisher information}. $J$ can be interpreted as the
intrinsic size of $\Omega$ \cite{Gruenwald:07book}. Although not
essential to this work, it will be instructive to occasionally plug
it into our expressions.
%
As distance we choose the Hellinger distance.

\begin{theorem}[$\LossT_h^m(\t,D)$ for large $m$]\label{thmLTEmggn}
Under Assumption \ref{assmggn}, for point estimation, the predictive
Hellinger loss for large $m$ is
\bqan
  \LossT_h^m(\t,D) &=& 2-2\left({8\pi\over m}\right)^{d/2}
  {p(\t|D)\over\sqrt{\det I_1(\t)}}[1+O(m^{-1/2})]
\\
  &\stackrel{J}=& 2-2\left({8\pi\over m}\right)^{d/2}
  {p(D|\t)\over J p(D)}[1+O(m^{-1/2})]
\eqan
where the first expression holds for any continuous prior density
and the second expression ($\stackrel{J}=$) holds for Jeffrey's
prior.
\end{theorem}

\paradot{IMAP}
The asymptotic expression shows that minimizing $\LossT_h^m$ is
equivalent to the following maximization
\beq\label{ipost}
   \IMAP:\qquad
   \tilde\t_h^\infty \;=\; \t^\IMAP \;:=\; \arg\max_\t{p(\t|D)\over\sqrt{\det I_1(\t)}}
\eeq
Without the denominator, this would just be MAP estimation. We have
discussed that MAP is not reparametrization invariant, hence
can be corrupted by a bad choice of parametrization. Since the
square root of the Fisher information transforms like the posterior,
their ratio is invariant. So PHI led us to a nice reparametrization
invariant variation of MAP, immune to this problem.
Invariance of the expressions in Theorem \ref{thmLTEmggn} is not a
coincidence. It has to hold due to Proposition \ref{propI}.
For Jeffrey's prior (second expression in Theorem \ref{thmLTEmggn}),
minimizing $\LossT_h^m$ is equivalent to maximizing the likelihood,
i.e.\ {$\tilde\t_h^\infty=\t^\ML$}.
Remember that the expressions are exact even and especially for
small samples $D_n$. No large $n$ approximation has been made. For
small $n$, MAP, ML, and IMAP can lead to significantly different
results. For Jeffrey's prior, IMAP and ML coincide. This is a nice
reconciliation of MAP and ML: An ``improved'' MAP leads for
Jeffrey's prior back to ``simple'' ML.

\paradot{MDL}
We can also relate PHI to MDL by taking the logarithm of the second expression
in Theorem \ref{thmLTEmggn}:
\beq\label{PHIMDL}
  \tilde\t_h^\infty \;\stackrel{J}=\; \arg\min_\t\{-\log p(D|\t) + \fr d2\log\fr{m}{8\pi} + J\,\}
\eeq
For $m=4n$ this is the classical (large $n$ approximation of) MDL
\cite{Gruenwald:07book}. So presuming that \req{PHIMDL} is a
reasonable approximation of PHI even for $m=4n$, MDL approximately
minimizes the predictive Hellinger loss {\em iff} used for $O(n)$
predictions. We will not expand on this, since the alluded
relation to MDL stands on shaky grounds (for several reasons).

\begin{corollary}[$\tilde\t_h^\infty=\t^\IMAP\stackrel{J}=\t^\ML$]\label{corLTEmggn}
The predictive estimator $\tilde\t_h^\infty=\lim_{m\to\infty}\arg\min_\t
\LossT_h^m(\t,D)$ coincides with $\t^\IMAP$, a representation
invariant variation of MAP. In the special case of Jeffrey's prior, it
also coincides with the maximum likelihood estimator $\t^\ML$.
\end{corollary}

\begin{theorem}[$\LossT_h^m(\T,D)$ for large $m$]\label{thmLTCmggn}
Under Assumption \ref{assmggn}, for composite $\T$, the predictive
Hellinger loss for large $m$ is
\bqan
  \LossT_h^m(\T,D) &=& 2-2\left({8\pi\over m}\right)^{d/4}
  {1\over\sqrt{\P[\T]}}\int_\T p(\t|D)\sqrt{p(\t)\over\sqrt{\det I_1(\t)}}d\t+o(m^{-d/4})
\\
  &\stackrel{J}=& 2-2\left({8\pi\over m}\right)^{d/4}
  \sqrt{p(D|\T)\P[\T|D]\over J\P[D]}+o(m^{-d/4})
\eqan
where the first expression holds for any continuous prior density
and the second expression ($\stackrel{J}=$) holds for Jeffrey's
prior.
\end{theorem}

\paradot{MAP meets ML half way}
The second expression in Theorem \ref{thmLTCmggn} is proportional to
the geometric average of the posterior and the composite likelihood.
For large $\T$ the likelihood gets small, since the average involves
many wrong models. For small $\T$, the posterior is proportional to
the volume of $\T$ hence tends to zero. The product is maximal for
some $\T$ in-between:
\beq\label{mlmap}\textstyle
  \ML\!\times\!\MAP:\;
  \sqrt{p(D|\T)\P[\T|D]\over \P[D]}
  = {\P[\T|D]\over\sqrt{P[\T]}}
  = {p(D|\T)\sqrt{P[\T]}\over P[D]} \rightarrow
  {\scriptsize\left\{\begin{array}{ccl}
    1 & \mbox{for} & \T\to\Omega \\
    0 & \mbox{for} & \T\to\{\t\} \\
    O(n^{d/4}) & \mbox{for} & |\T|\sim n^{-d/2}
  \end{array}\right.}
\eeq
The regions where the posterior density $p(\t|D)$ and where the
(point) likelihood $p(D|\t)$ are large are quite similar, as long as
the prior is not extreme. Let $\T_0$ be this region. It typically
has diameter $O(n^{-1/2})$. Increasing $\T\supset\T_0$ cannot
significantly increase $\P[\T|D]\leq 1$, but significantly decreases
the likelihood, hence the product gets smaller. Vice versa,
decreasing $\T\subset\T_0$ cannot significantly increase
$p(D|\T)\leq p(D|\t^\ML)$, but significantly decreases the
posterior. The value at $\T_0$ follows from
$\P[\T_0]\approx\mbox{Volume}(\T_0) \approx O(n^{-d/2})$.
Together this shows that $\T_0$ approximately maximizes
the product of likelihood and posterior.
So the best predictive $\T_0=\tilde\T_h^\infty$ has diameter
$O(n^{-1/2})$, which is a very reasonable answer. It covers well but
not excessively the high posterior and high likelihood regions
(provided $\H$ is sufficiently rich of course). By multiplying the
likelihood or dividing the posterior with only the square root of
the prior, they meet half way!

\paradot{Bernoulli Example}
A Bernoulli process with uniform prior and $n_0=n_1$ has posterior
variance $\sigma_n^2={1\over 4n}$. Hence any reasonable symmetric
interval estimate $\T=[\fr12-z;\fr12+z]$ of $\t$ will have size
$2z=O(n^{-1/2})$. For PHI we get
\beqn
  {\P[\T|D]\over\sqrt{\P[\T]}}
  \;=\; {1\over\sqrt{2z}}{(n\!+\!1)!\over n_1!n_0!}\int_\T\t^{n_1}(1-\t)^{n_0}d\t
  \;\simeq\; {1\over\sqrt{2z}}\text{erf}\Big({z\over\sigma_n\sqrt{2}}\Big)
\eeqn
where equality $\simeq$ is a large $n$ approximation, and
erf$(\cdot)$ is the error function \cite{Abramowitz:74}.
erf$(x)/\sqrt{x}$ has a global maximum at $x\doteq 1$ within 1\%
precision. Hence PHI selects an interval of half-width $z\doteq
\sqrt{2}\sigma_n$.

If faced with a binary decision between point estimate
$\T_f=\{\fr12\}$ and vacuous estimate $\T_v=[0;1]$, comparing the
losses in Theorems \ref{thmLTEmggn} and \ref{thmLTCmggn}, we see
that for large $m$, $\T_v$ is selected, despite $\sigma_n$ being
close to zero for large $n$. In Section \ref{secPrelim} we have
explained that this makes from a predictive point of view. %
\eoe

Finally note that \req{mlmap} does not converge to (any monotone
function of) \req{ipost} for $\T\to\{\t\}$, i.e.\ and
$\tilde\T_h^\infty \not\rightarrow \tilde\t_h^\infty$, since
the limits $m\to\infty$ and $\T\to\{\t\}$ do not exchange.

\paradot{Finding $\tilde\T_h^\infty$}
Contrary to MAP and ML, an unrestricted maximization of \req{mlmap}
over {\em all} measurable $\T\subseteq\Omega$ makes sense. The
following result reduces the optimization problem to finding the
level sets of the likelihood function and to a one-dimensional
maximization problem.

\begin{theorem}[Finding $\tilde\T_h^\infty=\T^{\ML\!\times\MAP}$]\label{thmMLMAPfind}
Let $\T_\g:=\{\t:p(D|\t)\geq\g\}$ be the $\g$-level set of $p(D|\t)$.
If $\P[\T_\g]$ is continuous in $\g$, then
\beqn
  \T^{\ML\!\times\MAP} \;:=\; \arg\max_\T {\P[\T|D]\over\sqrt{\P[\T]}}
  \;=\; \mathop{\arg\max}_{\T_\g:\g\geq 0} {\P[\T_\g|D]\over\sqrt{\P[\T_\g]}}
\eeqn
More precisely, every global maximum of \req{mlmap} differs from
the maximizer $\T_\g$ at most on a set of measure zero.
\end{theorem}

Using posterior level sets, i.e.\ shortest $\alpha$-credible
sets/intervals instead of likelihood level sets would not work
(an indirect proof is that they are not RI).
%
For a general prior, $p(D|\t)\sqrt{p(\t)/I_1(\t)}$ level sets
need to be considered.
%
The continuity assumption on $\P[\T_\g]$ excludes likelihoods with
plateaus, which is restrictive if considering non-analytic
likelihoods. The assumption can be lifted by considering all $\T_\g$
in-between $\T^o_\g:=\{\t:p(D|\t)>\g\}$ and
$\bar\T_\g:=\{\t:p(D|\t)\geq\g\}$. Exploiting the special form of
\req{mlmap} one can show that the maximum is attained for either
$\T^o_\g$  or $\bar\T_\g$ with $\g$ obtained as in the theorem.

\paradot{Large $n$}
For large $n$ ($m\gg n\gg 1$), the likelihood usually tends to an
(un-normalized) Gaussian with mean=mode $\bar\t=\t^\ML$ and
covariance matrix $[n I_1(\bar\t)]^{-1}$. Therefore the levels sets
are ellipsoids
\beqn
  \T_r \;=\; \{\t:(\t-\bar\t)^\trp I_1(\bar\t)(\t-\bar\t)\leq r^2\}
\eeqn
We know that the size $r$ of the maximizing ellipsoid scales with
$O(n^{-1/2})$. For such tiny ellipsoids, \req{mlmap} is
asymptotically proportional to
\beqn
  {\P[\T_r|D]\over\sqrt{\P[\T_r]}}
  \;\propto\; {\int_{\T_r} p(D|\t)d\t\over\sqrt{\mbox{Volume}[\T_r]}}
  \;\propto\; {\int_{||z||\leq\rho}\e^{-||z||^2/2}dz\over\sqrt{\int_{||z||\leq\rho}1dz}}
  \;\propto\; {\int_0^{\rho^2/2} t^{d/2-1}\e^{-t}dt\over \rho^{d/2}}
  \;=\; {\gamma(\frs{d}2,\!\frs{\rho^2\!}2)\over \rho^{d/2}}
\eeqn
where $z:=\sqrt{n I_1(\bar\t)}(\t-\bar\t)\in\SetR^d$, and
$\rho:=r\sqrt n$, and $t:=\fr12\rho^2$, and $\gamma(\cdot,\cdot)$ is
the incomplete Gamma function \cite{Abramowitz:74}, and we dropped
all factors that are independent of $r$. The expressions also holds
for general prior in Theorem \ref{thmLTEmggn}, since asymptotically
the prior has no influence. They are maximized for the following
$\tilde r$:
\beqn\small
  \begin{array}{c||c|c|c|c|c|c|c|c|c}
    d                  &  1  &  2  &  3  &  4  &  5  & 10  & 100 & \cdots& \infty \\ \hline
    \tilde r\sqrt{n/d} &1.400&1.121&1.009&0.947&0.907&0.819&0.721& \cdots& 1/\sqrt{2} \\
  \end{array}
\eeqn
i.e.\ for $m\gg n\gg 1$, unrestricted PHI selects ellipsoid
$\tilde\T_h^\infty=\T_{\tilde r}$ of (linear) size $O(\sqrt{d/n})$.

So far we have considered $\LossT_h^m$. Analogous asymptotic
expressions can be derived for $\LossP_h^m$:
While $\LossP_h^m$ differs from $\LossT_h^m$, for point estimation their
minima {\boldmath$\hat\t_d^\infty=\tilde\t_d^\infty=\t^\IMAP$} coincide.
For composite $\T$, the answer is qualitatively similar but differs
quantitatively.

\section{Large Sample Approximations}\label{secmlln}

In this section we will study PHI for large sample sizes $n$,
more precisely the $n\gg m$ regime. For simplicity we concentrate on
the univariate $d=1$ case only. Data may be non-i.i.d.

\paradot{Sequential moment fitting (SMF)}
A classical approximation of the posterior density $p(\t|D)$ is by a
Gaussian with same mean and variance. In case the class of available
distributions is further restricted, it is still reasonable to
approximate the posterior by the distribution whose mean and
variance are closest to that of $p(\t|D)$. There might be a
tradeoff between taking a distribution with good mean (low bias) or
one with good variance. Often low bias is of primary importance, and
variance comes second. This suggests to first fit the mean, then the
variance, and possibly continue with higher order moments.

PHI is concerned with predictive performance, not with density
estimation, but of course they are related. Good density estimation
in general and sequential moment fitting (SMF) in particular lead
to good predictions, but the converse is not necessarily true.
We will indeed see that PHI for $n\to\infty$ (under certain
conditions) reduces to an SMF procedure.

\paradot{The SMF algorithm}
In our case, the set of available distributions is given by
$\{p(\t|\T):\T\in\H\}$. For some event $A$, let
\beq\label{moments}
  \bar\t^A := \E[\t|A] = \int\,\t\; p(\t|A)d\t \qmbox{and}
  \mu_k^A :=\E[(\t-\bar\t^A)^2|A] \quad (k\geq 2)
\eeq
be the mean and central moments of $p(\t|A)$. The posterior moments
$\mu_k^D$ are known and can in principle be computed. SMF
sequentially ``fits'' $\mu_k^\T$ to $\mu_k^D$:
Starting with $\H_0:=\H$, let $\H_k\subseteq \H_{k-1}$ be the
set of $\T\in\H_{k-1}$ that minimize $|\mu_k^\T-\mu_k^D|$:
\beqn
  \H_k:=\{\mathop{\arg\min}_{\;\;\T\in\H_{k-1}}|\mu_k^\T-\mu_k^D|\},
  \qquad \H_0:=\H,\qquad \mu_1^A := \bar\t^A
\eeqn
Let $k^*:=\min\{k:\mu_k^\T\neq\mu_k^D,\T\in\H_k\}$ be the smallest
$k$ for which there is no perfect fit anymore (or $\infty$
otherwise). Under some quite general conditions, in a certain sense,
all and only the $\T\in\H_{k^*}$ minimize $\LossP_d^m(\T,D_n)$ for
large $n$.

\begin{theorem}[PHI for large $n$ by SMF]\label{thmPHISMF}
For some $k\leq k^*$, assume $p(\vx|\t)$ is $k$ times continuously
differentiable w.r.t.\ $\t$ at the posterior mean $\bar\t^D$. Let
$\beta>0$ and assume $\sup_\t\int|p^{(k)}(\vx|\t)|^\beta d\t<\infty$,
$\mu_k^D=O(n^{-k/2})$, $\mu_k^\T=O(n^{-k/2})$,
and $d(p,q)/|p-q|^\beta$ is a bounded function. Then
\beqn
  \LossP_d^m(\T,D) \;=\; O(n^{-k\beta/2})
  \quad\forall\T\in\H_k \quad (k\leq k^*)
\eeqn
\end{theorem}

For the $\a\leq 1$ distances we have $\beta=1$, for the square
distance we have $\beta=2$ (see Section \ref{secPHIP}). For i.i.d.\
distributions with finite moments, the assumption
$\mu_k^D=O(n^{-k/2})$ is virtually nil.
%
Normally, no $\T\in\H$ has better loss order than
$O(n^{-{k^*}\beta/2})$, i.e.\ $\H_{k^*}$ can be regarded as the set
of all asymptotically optimal predictors. In many cases, $\H_{k^*}$
contains only a single element.
%
Note that $\H_{k^*}$ does neither depend on $m$, nor on the chosen
distance $d$, i.e.\ the best predictive hypothesis
$\hat\T=\hat\T_d^m$ is essentially the same for all $m$ and $d$ if
$n$ is large.

\paradot{Bernoulli Example}
In the Bernoulli Example in Section \ref{secPrelim}
we considered a binary decision between point estimate
$\T_f=\{\fr12\}$ and vacuous estimate $\T_v=[0;1]$, i.e.\
$\H_0=\{\T_f,\T_v\}$. For $n_0=n_1$ we have
$\bar\t^{[0;1]}=\bar\t^{\smash{\frs12}}=\fr12=\bar\t^D$, i.e.\ both fit the
first moment exactly, hence $\H_1=\H_0$. For the second moments we
have $\mu_2^D=\fr1{4n}$, but $\mu_2^{[0;1]}=\fr1{12}$ and
$\mu_2^{\smash{\frs12}}=0$, hence for large $n$ the point estimate matches
the posterior variance better, so
$\hat\T=\{\fr12\}\in\H_2=\{\T_f\}$, which makes sense.
\eoe

For unrestricted (single) point estimation, i.e.\
$\H=\{\{\t\},\t\in\SetR\}$, one can typically estimate the mean
exactly but no higher moments. More generally, finite mixture models
$\T=\{\t_1,...,\t_l\}$ with $l$ components (degree of freedoms) can
fit at most $l$ moments.
For large $l$, the number of $\t_i\in\hat\T$ that lie in a small
neighborhood of some $\t$ (i.e.\ the ``density'' of points in
$\hat\T$ at $\t$) will be proportional to the likelihood $p(D|\t)$.
%
Countably infinite and even more so continuous models if otherwise
unrestricted are sufficient to get all moments right. If the
parameter range is restricted, anything can happen ($k^*=\infty$ or
$k^*<\infty$).
%
For interval estimation $\H=\{[a;b]:a,b\in\SetR,a\leq b\}$ and uniform
prior, we have $\bar\t^{[a;b]}=\fr12(a+b)$ and $\mu_2^{[a;b]}={1\over
12}(b-a)^2$, hence the first two moments can be fitted exactly
and the SMF algorithm yields the unique asymptotic solution
$\hat\T=[\bar\t^D-\sqrt{3}\mu_2^D\,;\,\bar\t^D+\sqrt{3}\mu_2^D]$.
%
In higher dimensions, common choices of $\H$ are convex sets,
ellipsoids, and hypercubes. For ellipsoids, the mean and covariance
matrix can be fitted exactly and uniquely similarly to 1d interval
estimation.
%
While SMF can be continued beyond $k^*$, $\H_k$ typically does {\em not}
contain $\hat\T$ for $k>k^*$ anymore. The correct continuation
beyond $k^*$ is either $\H_{k+1}=\{\arg\min_{\T\in\H_k}\mu_k^\T\}$
or $\H_{k+1}=\{\arg\max_{\T\in\H_k}\mu_k^\T\}$ (there is some
criterion for the choice), but apart from exotic situations this
does not improve the order $O(n^{-k^*\beta/2})$ of the loss, and
usually $|\H_{k^*}|=1$ anyway.

Exploiting Theorem \ref{thmPreqTr}, we see that SMF is also
applicable for $\LossT_2^m$ and $\LossT_r^m$.
Luckily, Offline $\PHIT$ can also be reduced to 1-Batch $\PHIT$:

\begin{proposition}[Offline = 1-Batch]
If $x_{i\in\SetN}$ are i.i.d., the Offine $\LossT$ is proportional to the 1-Batch $\LossT$:
\beqn
  \LossT_d^{1m}(\T,D) \\
  \;:= \sum_{k=n}^{n+m-1}\int\LossT_d^1(\T,D_k)\,p(x_{n+1:k}|D)dx_{n+1:k}
  \;=\; m\,\LossT_d^1(\T,D)
\eeqn
In particular, Offine $\PHIT$ equals 1-Batch $\PHIT$:
$\tilde\T_d^{1m}=\tilde\T_d^1$.
\end{proposition}
Exploiting Theorem \ref{thmPreqTr}, we see that also
{\boldmath$\LossP_{2|r}^{1m}=m\LossP_{2|r}^m$}+constant.
Hence we can apply SMF also for Offline
$\PHIP_{2|r}$ and $\smash{\PHIT_{2|r}}$.
For square loss, i.i.d.\ is not
essential, independence is sufficient.

\section{Discussion}\label{secDisc}

\paradot{Summary}
If prediction is the goal, but full Bayes not feasible, one should
{\em identify} (estimate/test/select) the {\em hypothesis}
(parameter/model/interval) that {\em predicts} best. What best is can
depend on the problem setup: What our benchmark is ($\LossP$,
$\LossT$), the distance function we use for comparison ($d$), how
long we use the model ($m$) compared to how much data we have at
hand ($n$), and whether we continue to learn or not (Batch,Offline).
We have defined some reparametrization and representation invariant
losses that cover many practical scenarios. Predictive hypothesis
identification (PHI) aims at minimizing this loss.
For $m\to\infty$,
PHI overcomes some problems of and even reconciles (a variation of)
MAP and (composite) ML. Asymptotically, for $n\to\infty$, PHI
reduces to a sequential moment fitting (SMF) procedure, which is
independent of $m$ and $d$.
The primary purpose of the asymptotic approximations was to gain
understanding (e.g.\ consistency of PHI follows from it), without
supposing that they are the most relevant in practice. A case where
PHI can be evaluated efficiently exactly is when a sufficient
statistic is available.

\paradot{Outlook}
There are many open ends and possible extensions that deserve
further study. Some results have only been proven for specific
distance functions. For instance, we conjecture that PHI reduces to
IMAP for other $d$ (seems true for $\a$-distances).
Definitely the behavior of PHI should next be studied for
semi-parametric models and compared to existing model (complexity)
selectors like AIC, LoRP \cite{Hutter:07lorp}, BIC, and MDL
\cite{Gruenwald:07book}, and cross validation in the supervised
case.
Another important generalization to be done is
to supervised learning (classification and regression), which
(likely) requires a stochastic model of the input variables.
PHI could also be generalized to predictive density estimation
proper by replacing $p(\vx|\T)$ with a (parametric) class of
densities $q_\vartheta(\vx)$.
Finally, we could also go the full way to a decision-theoretic setup
and loss.
Note that Theorems \ref{thmLTEmggn} and \ref{thmPHISMF} combined
with (asymptotic) frequentist properties like consistency of
MAP/ML/SMF easily yields analogous results for PHI.

\paradot{Conclusion}
We have shown that predictive hypothesis identification
scores well on all desirable properties listed in Section \ref{secPHIP}.
In particular, PHI can properly deal with nested hypotheses, and
nicely justifies, reconciles, and blends MAP and ML for $m\gg
n$, MDL for $m\approx n$, and SMF for $n\gg m$.

\paradot{Acknowledgements}
Many thanks to Jan Poland for his help improving the clarity
of the presentation.


\begin{small}

\end{small}


\begin{thebibliography}{MGB05}\parskip=0ex

\bibitem[AS74]{Abramowitz:74}
M.~Abramowitz and I.~A. Stegun, editors.
\newblock {\em Handbook of Mathematical Functions}.
\newblock Dover publications, 1974.

\bibitem[BB04]{Barbieri:04}
M.~M. Barbieri and J.~O. Berger.
\newblock Optimal predictive model selection.
\newblock {\em Annals of Statistics}, 32(3):870–--897, 2004.

\bibitem[BCH93]{Barron:93}
A.~R. Barron, B.~S. Clarke, and D.~Haussler.
\newblock Information bounds for the risk of {Bayesian} predictions and the
  redundancy of universal codes.
\newblock In {\em Proc. IEEE International Symposium on Information Theory
  (ISIT)}, pages 54--54, 1993.

\bibitem[Bis06]{Bishop:06}
C.~M. Bishop.
\newblock {\em Pattern Recognition and Machine Learning}.
\newblock Springer, 2006.

\bibitem[BM98]{Borovkov:98}
A.~A. Borovkov and A.~Moullagaliev.
\newblock {\em Mathematical Statistics}.
\newblock Gordon \& Breach, 1998.

\bibitem[Gei93]{Geisser:93}
S.~Geisser.
\newblock {\em Predictive Inference}.
\newblock Chapman \& Hall/CRC, 1993.

\bibitem[Gr{\"u}07]{Gruenwald:07book}
P.~D. Gr{\"u}nwald.
\newblock {\em The Minimum Description Length Principle}.
\newblock The MIT Press, Cambridge, 2007.

\bibitem[Hut05]{Hutter:04uaibook}
M.~Hutter.
\newblock {\em Universal Artificial Intelligence: Sequential Decisions based on
  Algorithmic Probability}.
\newblock Springer, Berlin, 2005.
\newblock 300 pages, http://www.hutter1.net/ai/uaibook.htm.

\bibitem[Hut07]{Hutter:07lorp}
M.~Hutter.
\newblock The loss rank principle for model selection.
\newblock In {\em Proc. 20th Annual Conf. on Learning Theory ({COLT'07})},
  volume 4539 of {\em LNAI}, pages 589--603, San Diego, 2007. Springer, Berlin.

\bibitem[KW96]{Kass:96}
R.~E. Kass and L.~Wasserman.
\newblock The selection of prior distributions by formal rules.
\newblock {\em Journal of the American Statistical Association},
  91(435):1343--1370, 1996.

\bibitem[Lan02]{Langford:02}
J.~Langford.
\newblock Combining train set and test set bounds.
\newblock In {\em Proc. 19th International Conf. on Machine Learning
  (ICML-2002)}, pages 331--338. Elsevier, 2002.

\bibitem[LF82]{Lejeune:82}
M.~Lejeune and G.~D. Faulkenberry.
\newblock A simple predictive density function.
\newblock {\em Journal of the American Statistical Association},
  77(379):654--657, 1982.

\bibitem[MGB05]{Mukho:05}
N.~Mukhopadhyaya, J.~K. Ghosh, and J.~O. Berger.
\newblock Some {B}ayesian predictive approaches to model selection.
\newblock {\em Statistics \& Probability Letters}, 73(4):2005, 2005.

\bibitem[Wal96]{Walley:96}
P.~Walley.
\newblock Inferences from multinomial data: learning about a bag of marbles.
\newblock {\em Journal of the Royal Statistical Society B}, 58(1):3--57, 1996.

\bibitem[Zuc00]{Zucchini:00}
W.~Zucchini.
\newblock An introduction to model selection.
\newblock {\em Journal of Mathematical Psychology}, 44(1):41--61, 2000.

\end{thebibliography}
\end{document}